\documentclass[3p, twocolumn]{elsarticle}
\usepackage{graphicx}
\usepackage{amsmath}
\usepackage{multirow}
\usepackage{subcaption}
\usepackage[linesnumbered,ruled,vlined]{algorithm2e}
\usepackage{float}
\usepackage{booktabs}
\usepackage{tikz}
\usepackage{pgfplots}
\pgfplotsset{compat=1.3}
\usepackage{subcaption}

\begin{document}

\begin{frontmatter}
\title{AILearn: An Adaptive Incremental Learning Model for\\ Spoof Fingerprint Detection}
\author[label1]{Shivang Agarwal}\ead{shivanga.rs.cse16@iitbhu.ac.in}
\author[label2]{Ajita Rattani}\ead{ajita.rattani@wichita.edu}
\author[label1]{C. Ravindranath Chowdary\corref{cor1}}\ead{rchowdray.cse@iitbhu.ac.in}\cortext[cor1]{Corresponding author}
\address[label1]{Department of Computer Science and Engineering,
			Indian Institute of Technology (BHU), Varanasi, India 221005}
\address[label2]{Department of Electrical Engineering and Computer Science, Wichita State University, USA }
\begin{abstract}
			Incremental learning enables the learner to accommodate new knowledge without retraining the existing model. It is a challenging task which requires learning from new data as well as preserving the knowledge extracted from the previously accessed data. This challenge is known as the stability-plasticity dilemma. We propose AILearn, a generic model for incremental learning which overcomes the stability-plasticity dilemma by carefully integrating the ensemble of base classifiers trained on new data with the current ensemble without retraining the model from scratch using entire data. We demonstrate the efficacy of the proposed AILearn model on spoof fingerprint detection application. One of the significant challenges associated with spoof fingerprint detection is the performance drop on spoofs generated using new fabrication materials. AILearn is an adaptive incremental learning model which adapts to the features of the ``live'' and ``spoof'' fingerprint images and efficiently recognizes the new spoof fingerprints as well as the known spoof fingerprints when the new data is available. To the best of our knowledge, AILearn is the first attempt in incremental learning algorithms that adapts to the properties of data for generating a diverse ensemble of base classifiers. From the experiments conducted on standard high-dimensional datasets LivDet 2011, LivDet 2013 and LivDet 2015, we show that the performance gain on new fake materials is significantly high. On an average, we achieve $49.57\%$ improvement in accuracy between the consecutive learning phases. 
		\end{abstract}
	\end{frontmatter}
\section{Introduction}
	Incremental learning is a process of learning in the presence of new data while retaining the knowledge acquired from previously seen data. Incremental learning is useful for applications that require accessing a huge amount of data in regular chunks because it does not need to retrain the model on the entire data when the model needs to expand progressively. For instance, in spoof fingerprint detection, where the task is to classify the fingerprint images as ``live" and ``spoof", the learning model is expected to learn incrementally from fingerprint images generated using novel fabrication materials. Therefore, the learning model must preserve the knowledge $H_t$ extracted from previously seen data $D_{Train_t}$ of live and spoof fingerprint images while learning from the upcoming data $D_{Train_{t+1}}$ without accessing $D_{Train_t}$. After learning from $D_{Train_{t+1}}$, the knowledge $H_{t+1}$ must be carefully integrated with $H_t$. Therefore,
	\begin{equation}
	H_{t+1} \leftarrow D_{Train_{t+1}} \cup H_t 
	\end{equation}\label{eq1}
	
	Polikar et al. \cite{983933} define a set of properties that an efficient incremental learning algorithm must possess: 1) it must learn new knowledge from upcoming data, 2) it should not require to access old training data, 3) it should preserve the previously acquired knowledge, and 4) it should be able to accommodate new concepts that may be available in the novel data. To address these properties, we propose AILearn, an adaptive incremental learning algorithm based on ensemble learning, which learns from the new data while retaining the previous knowledge without requiring to access the old data.
    
    The major challenge in incremental learning is to overcome the stability-plasticity dilemma, where stability signifies retaining the previously acquired knowledge, and plasticity signifies learning from new data \cite{983933}. Therefore, an ideal approach for incremental learning must find a balance between stability and plasticity. Focusing only on plasticity may lead to a situation, called \emph{catastrophic forgetting} where the learning model forgets the previously acquired knowledge while learning from new data \cite{FGCS-survey}. At the same time, concentrating only on stability may lead to the inability of capturing comprehensive knowledge from the latest data \cite{LI2020105694}.
    
    Another challenge in incremental learning is learning in the presence of concept drift \cite{5975223, 8726784, FGCS-CD}. Concept drift is a situation where the underlying data distribution changes over time, such that
\begin{equation}\label{eq10}    
p_{t+1}(x, w) \neq p_{t}(x, w)
\end{equation}
    where, $x$ represents an instance, $w$ is the class label associated with $x$, and $t$ is the timestamp.
    
    Incremental learning can be applied in various applications such as credit card fraud detection \cite{DALPOZZOLO20144915, Chan:1998:TSL:3000292.3000320}, object recognition in image processing \cite{4633694}, video surveillance in computer vision \cite{LU2014132}, interactive kinesthetic teaching in robotics \cite{DBLP:conf/icra/SaverianoAL15}, automated annotation for video and speech tagging \cite{DBLP:conf/esann/GepperthH16}, science article recommendation \cite{LUO2012271}, image recognition \cite{ROY2020148}, text classification \cite{SHAN2020113198}, object learning \cite{8953997}, social network analysis \cite{FGCS-Application-ii}, crypto-ransomware detection \cite{FGCS-application} and similar applications. It is particularly useful in applications where it is required to learn from data in multiple phases one chunk at a time or the applications where the size of data is so vast that it requires to be broken into numerous parts and accessed in multiple phases. In addition, incremental learning is also useful in applications where the old data is no longer available. Therefore the model cannot be retrained from the entire data; instead, it has to be incrementally updated.
    
    As single transaction information is typically not sufficient to mark a transaction as legitimate or fraudulent, incremental learning is used for the credit card fraud detection \cite{DALPOZZOLO20144915}. The fraud detector needs to be updated as soon as new data arrives. Updating the model with new knowledge is required because the usage behaviour of the cardholder /fraudster changes regularly. If the fraud evolves over time, the fraud-detection model must adapt to the new distribution. Incremental learning enables the model to deal with the changes in usage pattern in a non-stationary environment by retaining the past behaviour and learning from the new behaviour.
    
    Vehicle classification is another application where incremental learning plays a vital role \cite{WEN2015395}. In vehicle detection application, new image data is regularly added in batches over a period of time, and the model requires to learn the new knowledge incrementally. Standard classifiers such as AdaBoost, SVMs, etc. in their natural forms, are not capable of accommodating the knowledge extracted from new images while retaining the knowledge extracted from previously seen images, therefore require to retrain the model which is cost expensive.
    
    The Matrix-Factorization (MF) based models are widely used with collaborative filtering based recommender systems \cite{LUO2012271}. In real-world scenarios, user feedbacks are constantly generated. To accommodate the new responses, if the model has to be retrained, it causes a massive amount of repetitive training overhead. Therefore MF models need to be incrementally updated with new feedbacks. 
	
	Traditional classification models are ill-equipped for handling catastrophic forgetting while learning object recognition incrementally \cite{DBLP:conf/iccv/ShmelkovSA17}. Models for object recognition require to learn from new data when neither the original training data nor annotations for the original classes in the new training set are available.
	
	In multimedia information retrieval and computer vision applications, the most essential and tedious step is to reduce the human effort for generating ground truth from large scale visual data while not compromising with the quality of annotations \cite{BIANCO201588}. In order to improve the efficacy of automatic annotation, it is required to expand the domain knowledge regularly. Therefore, the object instances need to be added in the object database to increase the robustness of existing object detectors while learning from new object data. 
	
	The learning behaviour varies depending on the application to which it has been applied. Learning methods can be grouped into three categories based on applications with different data requirements:
	\begin{enumerate}
		\item Applications that require access to the previously seen data. Predicting the stock exchange is one of these applications where we need yesterday's data and today's data to predict for tomorrow \cite{CHEN2017340}.  
		\item Applications that need access only to the knowledge extracted from the previously accessed data but not the actual data itself. Cancer diagnosis belongs to such an application where we need only the knowledge extracted from the previous bunch of data along with the current data \cite{GU2018196}.
		\item Applications that require discarding the previous data and building a new model strictly based on the current data. Analysing the current trends on social media such as Twitter is one of these applications where the previously accessed data become insignificant over time, and the learning model must be built entirely on the current data \cite{Saha:2012:LEE:2124295.2124376}. 
	\end{enumerate}
	The difference in the nature of incremental learning and conventional single-phase learning can be understood by the above classification. A model following the incremental learning paradigm requires to learn from multiple chunks of data, extract the knowledge from them, discard the previously seen data and make use of the already acquired knowledge along with the current data. Another variant of learning paradigm is online learning which learns from each incoming instance. Due to its highly dynamic nature, online learning suffers from poor stability and high computational cost \cite{LI2020105694}. On the other hand, the conventional single-phase learning requires discarding the existing classifier whenever new data is available and retraining the model from the ground truth.  
	
\begin{figure*}
	\centering
	\resizebox{\textwidth}{!}{\begin{tabular}{cccccc}
			\subcaptionbox{Live}{\includegraphics[width = 1in]{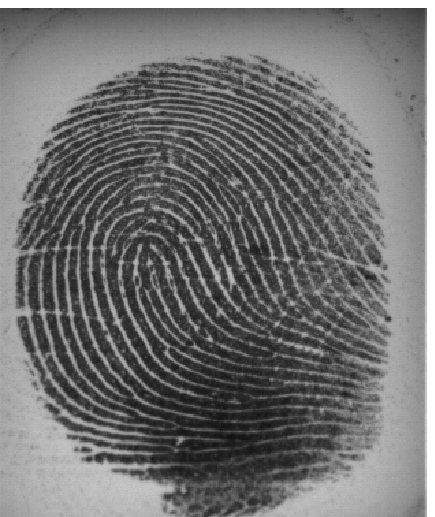}} &
			\subcaptionbox{Ecoflex}{\includegraphics[width = 1in]{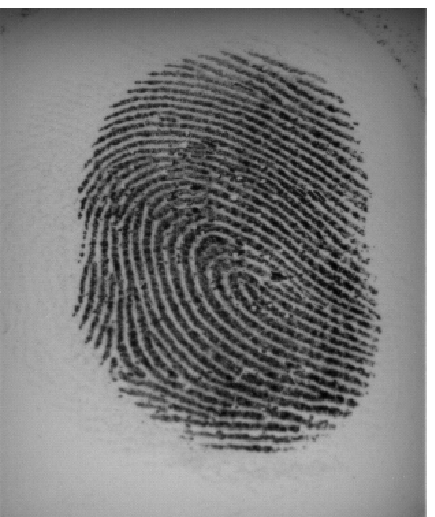}} &
			\subcaptionbox{Gelatin}{\includegraphics[width = 1in]{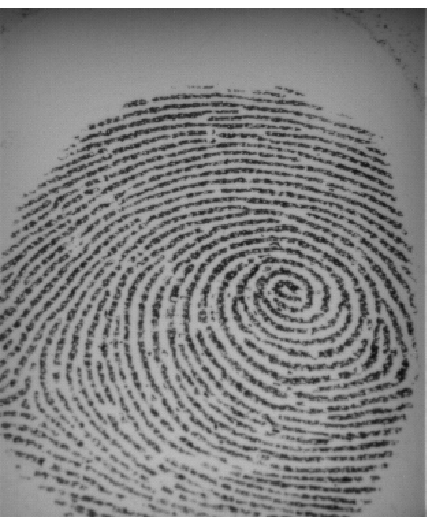}} &
			\subcaptionbox{Latex}{\includegraphics[width = 1in]{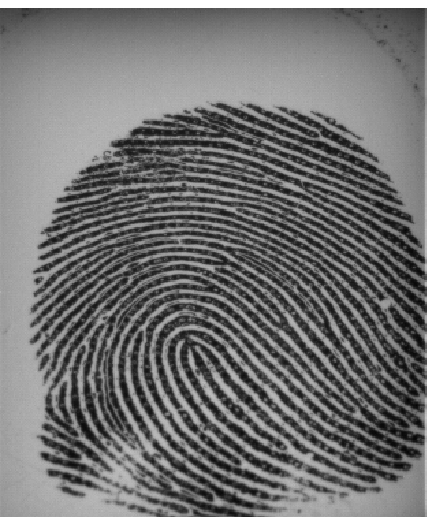}} &
			\subcaptionbox{Silgum}{\includegraphics[width = 1in]{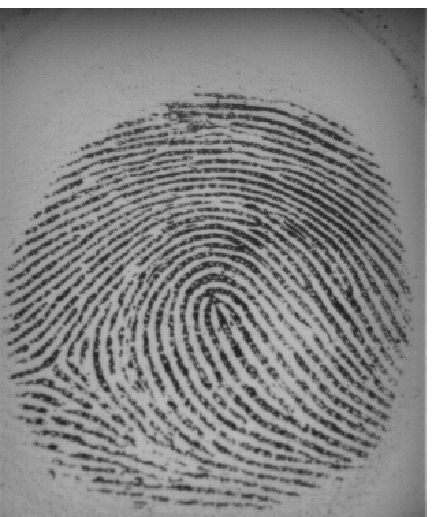}} &
			\subcaptionbox{WoodGlue}{\includegraphics[width = 1in, height = 1.2in]{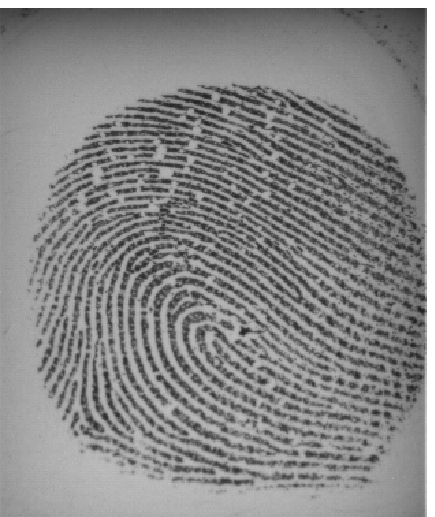}}\\
	\end{tabular}}
	\caption{Visual comparison between live and spoofs created using various spoof materials}
	\label{fig:spoof}
\end{figure*}

	In applications like fingerprint liveness detection, where the classifier learns from fingerprint images acquired from various sensors and generated using different fabrication materials, the properties of dataset play an essential role while learning. Existing studies suggest that fingerprint recognition systems are vulnerable to attacks by spoof fingerprints made of different materials such as gelatin, silicone, latex etc. \cite{JIA201491, 7029061}. As represented by Figure \ref{fig:spoof}, it is not possible for us to manually distinguish between live and spoof fingerprints generated using these materials. Further, the performance of the spoof fingerprint detector substantially degrades on the novel spoof materials \cite{KHO201952, 7180344}. Incremental learning is one of the solutions to mitigate performance degradation due to evolving spoofing techniques by using novel fabrication materials. Therefore, the spoof fingerprint detection application has been chosen for demonstrating the efficacy of incremental learning.  

In spoof fingerprint detection, our adaptive incremental learning model AILearn yields a robust spoof detector that can identify spoof fingerprints generated from fabrication materials unknown to the current model. We exploit the incremental learning scenario in spoof fingerprint detection by considering two learning phases. In the first phase, the model learns images from live category and spoof images of two fabrication materials. In the later phase, the model learns the spoof images of the remaining fabrication materials. The idea is to test the ability of the model to maintain stability and plasticity while obtaining performance gain on novel spoof materials in multiple phases. In an ideal scenario, the incremental learner must not observe a significant performance degradation on the known data while improving its performance on new data in subsequent learning phases.
	
	The contributions made by this study are as follows:
	\begin{enumerate}
		\item We propose a novel incremental learning model AILearn, which is adaptive towards the similarity inherently present in the data and is capable of overcoming the classic stability-plasticity dilemma. The proposed model produces lower performance degradation and higher performance improvement while learning in multiple phases.
		\begin{itemize}
			\item We perform clustering on the training data to generate clusters of instances and use these clusters to train RBF SVMs as base classifiers which results in an ensemble of diverse classifiers. In our observations, these base classifiers are free from catastrophic forgetting, i.e. while learning from new data, there is no significant performance loss concerning the previously trained instances.
		\end{itemize}
		\item Our proposed AILearn does not need to retrain the model from the scratch while introducing new data to it. As we use an ensemble of base classifiers, it offers us a high degree of reliability and robustness. The new knowledge is added by carefully integrating another ensemble to the model.	
		\item AILearn for spoof fingerprint detection does not need to access the previously seen fingerprints while learning the new fingerprints, which results in low memory requirements. In addition, it discards the poorly performing base classifiers and uses only the relevant ones to save the storage as well as improve the classification accuracy.  
	\end{enumerate}
%
	\section{Related Work}\label{related-work}
	\subsection{Incremental Learning}
%
	Incremental learning has benefited many applications that require learning in pieces. In \cite{DBLP:conf/comad/ChowdaryK08}, the authors propose a method for an incremental summary generation where the task is to update the current summary of documents after encountering a new document. Their proposed approach finds the most significant sentence in the document that can be replaced with a sentence in the current summary making it more accurate and updated. 
    
    Most of the approaches proposed for incremental learning so far make use of ensemble learning where a set of classifiers are learned from each chunk of training data. Learn++ \cite{983933} is a family of algorithms widely used for various applications involving incremental learning \cite{983933, 5975223, 4721600}. Learn++ algorithms address the issues like the stability-plasticity dilemma, catastrophic forgetting, concept drift etc. Learn++ was initially proposed for training neural network pattern classifiers incrementally. It generates weak hypotheses and combines them using a weighted majority voting scheme. Learn++.NSE \cite{5975223} enhances the previous version by accommodating concept drift in non-stationary environments. Learn++.NSE-SMOTE \cite{6235959} enhances the previous versions by overcoming the problem of class-imbalance in the streaming data. Irrespective of its popularity and broad applicability, Learn++ family of incremental learning models do not consider the properties of data while generating the base learners. 
    
    There are some cost-sensitive learning approaches proposed in the past that assign unequal misclassification costs to different classes \cite{DBLP:conf/ijcai/GuSL15}. Incremental learning becomes useful as it keeps track of the concept drift and manages it in a way that it does not increase false negatives. Another approach for detecting concept drift for incremental learning on non-stationary environments is proposed in \cite{8674766}. The authors claim that the early detection of concept drift may result in improved accuracy.  
    
    Recently, there has been some research on handling the class-imbalance while learning incrementally. In \cite{8954008}, the proposed learning model scales up to a large number of classes while managing the data imbalance between previously observed and new classes. While learning from the newly added data, it is essential to update the hypothesis accordingly. In \cite{DBLP:journals/corr/abs-1903-07864}, the old and new learning models are consolidated via a double distillation training objective. An unlabelled auxiliary data is exploited to consolidate the two models. In \cite{LI2020105694}, the authors propose a dynamically updated ensemble algorithm for dealing with class-imbalance and concept drift.   
    
    AILearn performs clustering to generate an ensemble of base classifiers in each learning phase. We claim that advantages of clustering are two-fold: we can have a disjoint ensemble of classifiers (a must-have a property in ensemble learning) and the base classifiers grasp the features of the data which is useful while classifying similar but unknown test instances.  
    \subsection{Spoof Fingerprint Detection}
    The application we consider in this paper is spoof fingerprint detection, which has its importance in forensics and information security. Much research has been done in the field of spoof fingerprint detection, but most of it does not consider it as an application of incremental learning where fingerprints with novel spoof materials are added to the model. Various single-phase spoof detectors have been proposed in the past \cite{7042825,10.1007/978-3-642-17955-6_21,Marasco:2014:SAS:2658850.2617756,7390065}. The performance of single-phase spoof detectors degrades drastically when new fingerprints generated from novel spoof materials are introduced. We claim that the incremental aspect of the fingerprint liveness detection must be studied so that, the spoof detectors can accommodate new spoof materials efficiently. The recent research on spoof fingerprint detection shows the incremental behaviour by using the existing learning algorithms based on ensemble learning such as Learn++.NC \cite{4721600, KHO201952}, whereas some of the researchers find a way to modify SVMs to accommodate new knowledge to the model \cite{7180344}. We claim that while learning the base classifiers of the ensemble, the properties intrinsic to the dataset must be considered to form clusters of instances. Therefore, we propose a new learning algorithm and use it for spoof fingerprint detection. 
    
    \emph{To the best of our knowledge, AILearn is the first incremental learning algorithm which considers adaptiveness towards the similarity present in the data to generate a diverse set of base classifiers and integrate the ensembles incrementally in subsequent learning phases.}
	\section{AILearn: A Generic Model for Incremental Learning}\label{ailearn}
	The generic model of AILearn for incremental learning is described in Algorithm \ref{Algo1}. 
	\begin{algorithm}[!t]
		\caption{Learning Incrementally using AILearn Algorithm}
		\label{Algo1}
		\KwData{training data $D_{Train}=<x_s, y_s>$, \\
			test data $D_{Test}= <x_t, y_t>,$\\
			validation data $D_{Valid}= <x_u, y_u>,$\\
			a clustering algorithm $C$,\\
			a classification algorithm $K$,\\
			number of base classifiers $n$,\\
			number of learning phases $p$.}
		\KwResult{ensemble classifier $Z_f$}
		partition $D_{Train}$ and $D_{Test}$ into $p$ parts each;\\
		$Z_0$= NULL;\\
		\For{i= 1 to p}{
			$\{c_1, c_2, .., c_n\} \leftarrow C(D_{Train_i})$\;
			\For{j=1 to n}
			{
				$k_j \leftarrow K(c_j)$\;
				Check the accuracy $a_j$ of $k_j$ on $D_{Valid}$\;
			}
			$Z_i \leftarrow \{k_1, k_2, ..,k_n\}_i$\;
			$Z_i \leftarrow Z_i + Z_{i-1}$\;
		}
		$ Z_f \leftarrow Z_i$\;	
	\end{algorithm}
This model is application-independent and can be applied to various applications. In this paper, we consider an application where we fix the number of base classifiers to be generated in every learning phase. Therefore we consider the number of base classifiers $n$ as an input to the algorithm. Also, the proposed algorithm is independent of the choice of clustering and classification algorithms. 

To show the incremental behaviour of AILearn, we need to partition the original training dataset $D_{Train}$ into $p$ parts to be accessed in $p$ learning phases (step 1 in Algorithm \ref{Algo1}). We partition the test dataset $D_{Test}$ as well accordingly. The target is to demonstrate the incremental behaviour by adding a new batch of training data in each learning phase and testing the performance of the learned model on $D_{Test_{i+1}}$ as well as on $D_{Test_i}$.

Later, we perform clustering on each of the partitioned training set $D_{Train_i}$ that yields a set of clusters ${c_1, c_2, .., c_n}$. Next, we train base classifiers on each of the generated clusters $c_j$ by using a classification algorithm $K$ (step 4-6 of the algorithm). Therefore, we generate an ensemble of base classifiers where the diversity among the base classifiers is high due to clustering. As pointed out by Polikar et. al.\cite{983933}, if each classifier is trained on a different subset of training data based on some distribution, then it is highly probable that a misclassified instance is classified correctly by another classifier. Therefore, it is always advantageous to have a diverse set of base classifiers. 

After learning in every phase, first, we test the accuracy of the ensemble of base classifiers generated in the current phase on validation data which was held-out from training (step 7 of the algorithm). The purpose of using validation data is to assign weights to the base classifiers based on their performance. We use Equations \ref{eq3} and \ref{eq4} to assign weights to the classifiers. 
\begin{equation}
\label{eq3}
w_{y}^{x}=\sum_{i=1}^{n}a_{iy}^{x} 
\end{equation}
where, $w_{y}^{x}$ is the total weight associated with the class label $y$ for an instance $x$. $n$ is the number of base classifiers and $a_{iy}^{x}$ is the accuracy of $i^{th}$ base classifier which has predicted the class label $y$ for instance $x$ on validation data.
The final class label $y_f^x$ determined by the weighted majority is given by Equation \ref{eq3}:
\begin{equation}
\label{eq4}
y_{f}^{x}=argmax_y(w_{y}^{x})
\end{equation}
We test the accuracy of the ensemble $Z_i$ generated in $i^{th}$ learning phase (step 8) by classifying the instances of test data using Equations \ref{eq3} and \ref{eq4}.
Based on a threshold on the performance, the poorly performing classifiers are discarded so that they do not participate in the voting process. Similarly, in the next phase, we generate another ensemble of base classifiers $Z_{i+1}$, trained on the newly added data $D_{Train_{i+1}}$. We test the performance of individual base classifiers on validation data and merge the qualifying base classifiers with the existing ensemble from the previous phase (step 9). All the qualifying classifiers are tested on the test data to demonstrate the performance improvement. The idea is to highlight that as the model proceeds to learn $D_{Train_{i+1}}$; its performance is increased as tested on the new data without considerable performance loss on $D_{Train_{i}}$. 
\section{AILearn for Spoof Fingerprint Detection}\label{ailearn-spoof}
The incremental ability of the proposed algorithm is achieved by considering multiple learning phases. The performance degradation for the previously known knowledge and the performance improvement for the newly learned data are observed while moving to subsequent learning phases. 

\begin{figure*}[!t]
		\includegraphics[scale=0.25, width=\textwidth] {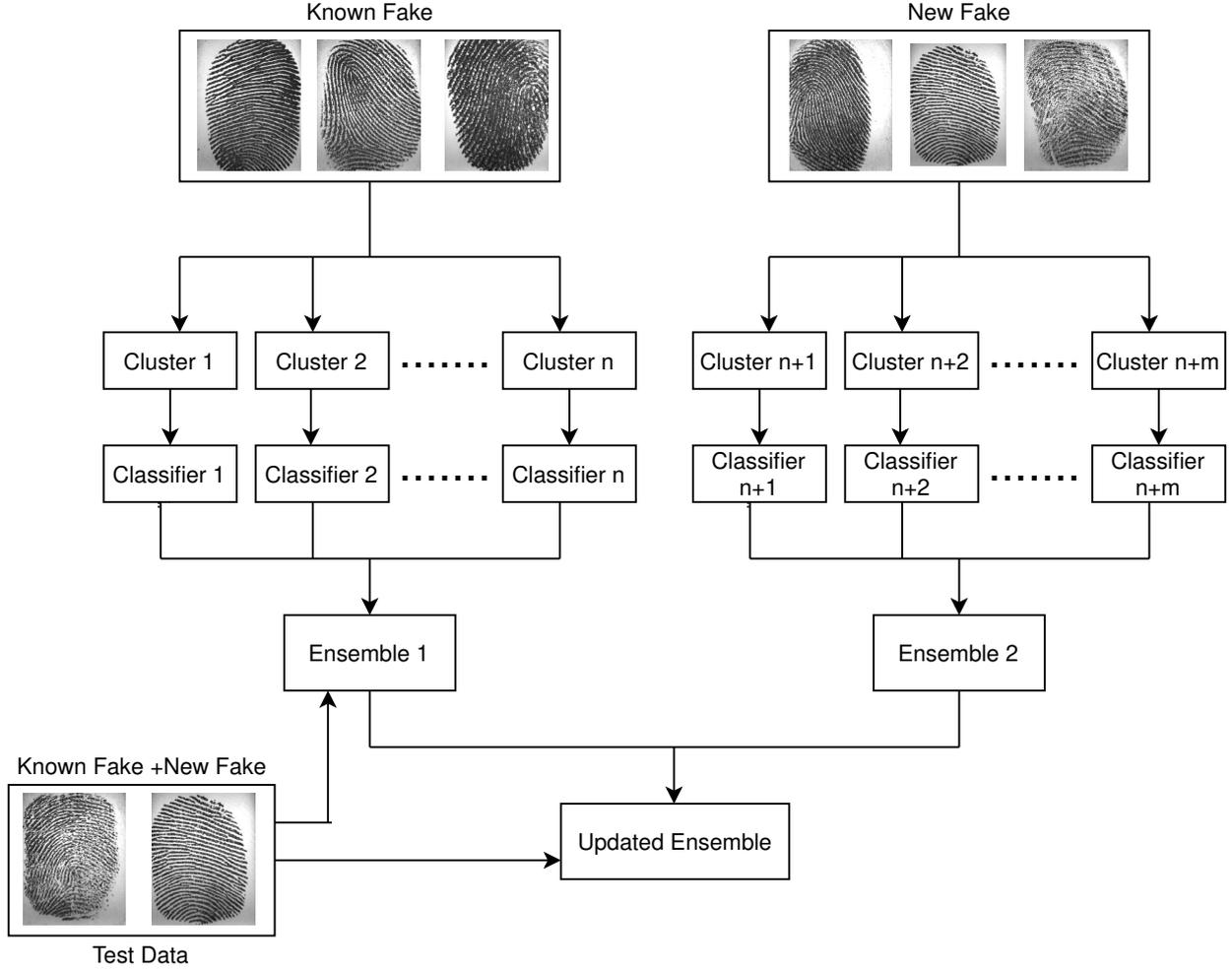}
		\caption{Schema of the proposed AILearn incremental learning algorithm for Spoof Fingerprint Detection.}
		\label{fig}
	\end{figure*}

The schematic diagram explaining the working mechanism of AILearn on spoof fingerprint detection is given in Figure \ref{fig}. In this study, we have used LivDet 2011 \cite{DBLP:conf/icb/YambayGDMRS12}, LivDet 2013 \cite{DBLP:conf/icb/GhianiYMTMRS13}, and LivDet 2015 \cite{DBLP:conf/btas/MuraGMRYS15} datasets. We partition each of the training data in two parts: Known Fake (KF) and New Fake (NF). KF consists of 1000 live instances and 400 spoof instances belonging to two of the spoof subcategories. NF consists of 600 spoof instances of the remaining subcategories. Test data is partitioned in the same manner (i.e., $Test_1$: $1000$ Live + $400$ Spoof, $Test_2$: $600$ Spoof). 

In the first phase, when the model is trained on KF, we test its accuracy on both $Test_1$ and $Test_2$. The idea is to see the difference in both accuracies as in the first phase, the model is trained only on KF, it must yield good accuracy on it but poor accuracy on NF. In the second phase, when new fake data is introduced to the training model, we need to accommodate it by managing two challenges: first, the knowledge acquired in the first phase must not be lost. Therefore, the accuracy of the updated model must not decrease drastically when tested on KF test data. This shows that the model possesses better stability. Second, the model must be able to learn from the new fake data added to the training model. Therefore the accuracy on NF must increase significantly, which shows that the model possesses better plasticity. 
The components of the proposed framework are explained as follows: 
\subsection{Feature Extraction}
\begin{enumerate}
\item Local Binary Patterns (LBP) are a local texture descriptor which is widely used for fingerprint liveness detection \cite{7390065,1717463}. LBP is an illumination invariant descriptor which determines the texture of an image by labelling each pixel with a binary value based on the thresholds on the neighbouring pixels. It considers the central pixel as the threshold and based on that it assigns the binary values to the neighbouring pixels. LBP value of the pixel is calculated by adding up the element-wise multiplications of the binary values with their weights. LBP histograms are robust in terms of grayscale variations, making them suitable for spoof fingerprint detection, as they can easily incorporate fingerprints with skin distortions, different skin qualities, and dry /moist /dirty skin. 
		
\item Local Phase Quantization (LPQ) can be an effective method for detecting liveness of a fingerprint image as it is insensitive to the blurring effect \cite{DBLP:conf/icpr/GhianiMR12,10.1007/978-3-540-69905-7_27}. LPQ features consider the spectral differences between live and spoof fingerprint images. 
		
\item Binarized Statistical Image Features (BSIF) is a method of constructing local image features to encode the texture information from the images \cite{DBLP:conf/icpr/KannalaR12}. The descriptors are determined by the statistical properties of natural image patches. For a particular image, BSIF computes a binary string for the pixels and use it as a local descriptor of the image intensity pattern in the pixel’s surroundings.
\item ResNet-50 \cite{DBLP:journals/corr/HeZRS15} is a deep Residual Network originally designed for object recognition. ResNet-50 has been pretrained on ImageNet database \cite{DBLP:journals/corr/RussakovskyDSKSMHKKBBF14}. By extracting the features using ResNet-50 we utilize transfer learning for spoof fingerprint detection. ResNet architecture is used for deep feature extraction as it is among the efficient Convolutional Neural Networks introduced till now. ResNet utilizes skip connections, or shortcuts to jump over layers to avoid the problem of vanishing gradients. 
\end{enumerate}
\subsection{Ensemble Generation}
After feature extraction, the proposed incremental learning algorithm generates the ensemble of base classifiers. The ensemble is created by using the following components:
    \begin{itemize}
        \item Training data $D_{Train} = {(x_1, y_1), ....(x_n, y_n)}$ contains training examples belonging to both ``live" and ``spoof" classes, where $x_i$ is a set of attributes generated from an image feature extraction algorithm and $y_i$ is the corresponding class label. 
        \item A clustering algorithm $C$ is used to cluster the training examples based on the similarity of records in the data. The target is to create a group of clusters {$c_1, .., c_n$} where the examples belonging to one cluster possess similar values of attributes whereas the examples belonging to different clusters possess different values of the attributes defined by image features. We strongly recommend using a clustering algorithm which does not require to define the number of clusters $n$ apriori, so that the clusters are naturally formed based on the similarities, but depending upon the application, the number of base clusters may be known apriori. 
        \item A classification algorithm $K$ to train the base classifiers on each $c_n$. $K$ uses each $c_n$ to generate a base classifier which is used to make a decision individually. In this way, the decision boundary of each base classifier is different from others, resulting in an ensemble of diverse base classifiers \cite{MELVILLE200599}. Later these decisions are integrated by using weighted majority voting to decide for the whole ensemble.
    \end{itemize} 
	
	\section{Experimental Results and Discussion}\label{experiment}
	\subsection{Dataset}\label{datasets}
	The description of the datasets used in this paper is given in Table \ref{dataset}. We use LivDet datasets of three years: LivDet 2011 \cite{DBLP:conf/icb/YambayGDMRS12}, LivDet 2013 \cite{DBLP:conf/icb/GhianiYMTMRS13} and LivDet 2015 \cite{DBLP:conf/btas/MuraGMRYS15}, which were used in fingerprint liveness detection competition conducted in consecutive years. The goal of this competition is to compare software-based fingerprint liveness detection methodologies and fingerprint systems which incorporate liveness detection capabilities. The datasets consist of fingerprint images broadly classified into two classes: ``live" and ``spoof". These fingerprints are tested on various biometric sensors: Biometrika, DigitalPersona, ItalData, Sagem etc. For all the sensors, we have approximately 1000 fingerprint images each for both of the classes in training and testing.
	\renewcommand{\arraystretch}{1.3}
	\begin{table*}[]
		\caption{The training and testing protocol of the experiments.}
		\begin{center}
			\scriptsize
			\resizebox{\textwidth}{!}{\begin{tabular}{llll}
					\hline
					Database   &                & Live (Train/Test) & Spoof (Train/Test)                                        \\ \hline
					LivDet2011 \cite{DBLP:conf/icb/YambayGDMRS12} & Biometrika     & 1000/1000        & 1000/1000 (ecoflex, gelatin, latex, silgum, wood glue)   \\ 
					& DigitalPersona & 1000/1000        & 1000/1000 (gelatin, latex, playdoh, silicone, wood glue) \\ 
					& ItalData       & 1000/1000        & 1000/1000 (ecoflex, gelatin, latex, silgum, wood glue)   \\ 
					& Sagem          & 1000/1000        & 1000/1000 (gelatin, latex, playdoh, silicone, wood glue) \\ \hline
					LivDet2013 \cite{DBLP:conf/icb/GhianiYMTMRS13} & Biometrika     & 1000/1000        & 1000/1000 (ecoflex, gelatin, latex, modasil, wood glue)   \\ 
					& ItalData       & 1000/1000        & 1000/1000 (ecoflex, gelatin, latex, modasil, wood glue)   \\ 
					  \hline
					LivDet2015 \cite{DBLP:conf/btas/MuraGMRYS15} & Biometrika     & 1000/1000        & 1000/1000 (ecoflex, gelatin, latex, wood glue)   \\ 
					& DigitalPersona       & 1000/1000        & 1000/1000 (ecoflex, gelatin, latex, wood glue)   \\ 
					  \hline  
			\end{tabular}}
		\end{center}
		\label{dataset}
	\end{table*}

Further, the images belonging to the spoof class can be categorized in multiple sub-categories based on the materials used for creating the spoof or fake fingerprint. These materials are gelatin, latex, playdoh, wood glue, silicone etc.  In LivDet 2011 \cite{DBLP:conf/icb/YambayGDMRS12} and LivDet 2013 \cite{DBLP:conf/icb/GhianiYMTMRS13}, each dataset has 200 images belonging to each of the five sub-categories, whereas in LivDet 2015 \cite{DBLP:conf/btas/MuraGMRYS15}, each dataset has 250 images belonging to each of the four sub-categories.
\subsection{Experimental settings}
As mentioned in the Section \ref{datasets}, we partition each of the training and testing data belonging to a particular sensor (e.g. Biometrika, DigitalPersona, etc.) into two parts: I. Known Fake (KF) and II. New Fake (NF). In LivDet 2011 \cite{DBLP:conf/icb/YambayGDMRS12} and LivDet 2013 \cite{DBLP:conf/icb/GhianiYMTMRS13}, we have five subcategories in the spoof class; therefore, we have ten possible combinations of training datasets for the known fake. In LivDet 2015 \cite{DBLP:conf/btas/MuraGMRYS15}, there are four sub-categories of spoof class, therefore we have six such combinations. In the first experimental setting, we partition the training data into two parts and learn each one of them in two learning phases (e.g., I. Live + Ecoflex + Gelatin, II. Latex + Silicone + Woodglue). The description of the partitioning of these datasets is given in Table \ref{partition}. In every phase, the trained model is tested on two test datasets created using the same setup. In the second experimental setting, we retrain the model in the second learning phase using the entire training data. We use the second experimental setting as the benchmark for the proposed model. Therefore, the motivation is to have competitive performance without the need for retraining the model using the entire data.
	\renewcommand{\arraystretch}{1.3}
	\begin{table*}[]
		\caption{Partitioning of the datasets in Phase I and Phase II for evaluation of the AILearn algorithm.}
		\begin{center}
			\resizebox{\textwidth}{!}
{\begin{tabular}{|l|l|l|l|l|l|l|}
\hline
\multirow{3}{*}{Sr. No.} & \multicolumn{2}{l|}{LivDet2011 \cite{DBLP:conf/icb/YambayGDMRS12}, LivDet2013 \cite{DBLP:conf/icb/GhianiYMTMRS13}}      & \multicolumn{2}{l|}{LivDet2011 \cite{DBLP:conf/icb/YambayGDMRS12}}                    & \multicolumn{2}{l|}{LivDet2015 \cite{DBLP:conf/btas/MuraGMRYS15}}          \\ \cline{2-7} 
                         & \multicolumn{2}{l|}{Biometrika, ItalData}        & \multicolumn{2}{l|}{Digital, Sagem}                & \multicolumn{2}{l|}{Biometrika, Digital} \\ \cline{2-7} 
                         & Phase I               & Phase II                 & Phase I                & Phase II                  & Phase I               & Phase II         \\ \hline
1                        & Live+Ecoflex+Gelatin  & Latex+Silgum+Woodglue    & Live+Gelatin+Latex     & Playdoh+Silicone+Woodglue & Live+Ecoflex+Gelatin  & Latex+Woodglue   \\ \hline
2                        & Live+Ecoflex+Latex    & Gelatin+Silgum+Woodglue  & Live+Gelatin+Playdoh   & Latex+Silicone+Woodglue   & Live+Ecoflex+Latex    & Gelatin+Woodglue \\ \hline
3                        & Live+Ecoflex+Silgum*  & Gelatin+Latex+Woodglue   & Live+Gelatin+Silicone  & Latex+Playdoh+Woodglue    & Live+Ecoflex+Woodglue & Gelatin+Latex    \\ \hline
4                        & Live+Ecoflex+Woodglue & Gelatin+Latex+Silgum     & Live+Gelatin+Woodglue  & Latex+Playdoh+Silicone    & Live+Gelatin+Latex    & Ecoflex+Woodglue \\ \hline
5                        & Live+Gelatin+Latex    & Ecoflex+Silgum+Woodglue  & Live+Latex+Playdoh     & Gelatin+Silicone+Woodglue & Live+Gelatin+Woodglue & Ecoflex+Latex    \\ \hline
6                        & Live+Gelatin+Silgum   & Ecoflex+Latex+Woodglue   & Live+Latex+Silicone    & Gelatin+Playdoh+Woodglue  & Live+Latex+Woodglue   & Ecoflex+Gelatin  \\ \hline
7                        & Live+Gelatin+Woodglue & Ecoflex+Latex+Silgum     & Live+Latex+Woodglue    & Gelatin+Playdoh+Silicone  &                       &                  \\ \hline
8                        & Live+Latex+Silgum     & Ecoflex+Gelatin+Woodglue & Live+Playdoh+Silicone  & Gelatin+Latex+Woodglue    &                       &                  \\ \hline
9                        & Live+Latex+Woodglue   & Ecoflex+Gelatin+Silgum   & Live+Playdoh+Woodglue  & Gelatin+Latex+Silicone    &                       &                  \\ \hline
10                       & Live+Silgum+Woodglue  & Ecoflex+Gelatin+Latex    & Live+Silicone+Woodglue & Gelatin+Latex+Playdoh     &                       &                  \\ \hline
\end{tabular}}
		\end{center}
		\label{partition}
	\end{table*}
	
This study can also be used for comparing the performance of spoof detectors with deep features and with hand-crafted local features. We use ResNet-50 \cite{DBLP:journals/corr/HeZRS15} for extracting deep features from the fingerprint images. LBP, LPQ and BSIF features were extracted using MATLAB. The extracted features are converted into arff files to make them WEKA compatible. We use Waikato Environment for Knowledge Analysis (Weka) \cite{Hall:2009:WDM:1656274.1656278} to classify the images into ``live" and ``spoof" classes. 
    
The proposed AILearn is not restricted to any particular clustering or classification algorithm. In our experiments, we use SimpleKMeans \cite{Arthur2007} clustering algorithm with $k$=2. We use SMO (John Platt's sequential minimal optimization algorithm for training a support vector classifier) \cite{Platt1998} as the classification algorithm. SVMs have been an appropriate choice for classifying fingerprint images as live and spoof \cite{AGARWAL2020113160, KHO201952, 7180344}. For every dataset, we report the overall accuracy of the model on known fake (KF) and new fake (NF) data as well as the bonafide presentation classification error rate (BPCER) and attack presentation classification error rate (APCER), where bonafide presentation = ``live", and attack presentation = ``spoof".      
\subsection{Results}\label{results}
In this section, we provide the experimental results conducted on three high dimensional datasets. To demonstrate the incremental behaviour of the proposed model, we learn the data in two phases. In the first phase, the model is trained over instances of ``Live'' category and instances belonging to two sub-categories of ``Spoof'' class. In the first phase, when the learned model is tested using the known sub-categories of spoof class, we call it ``I. Known Fake''. We also test the learned model on the remaining sub-categories on which the model is not yet trained, we call it ``I. New Fake''. In the second phase, we train the model on the remaining sub-categories of spoof fingerprints and integrate the hypotheses with the existing model. Now, again we test the performance of this updated model on the same test sets, we call them ``II. Known Fake'' and ``II. New Fake''. As AILearn does not require to access the whole training data in the second phase, the count of ``Live'' instances in the second training data is $0$; therefore in place of BPCER in New Fake, we have written N/A.     

Tables \ref{2011} and \ref{2013-15} describe the experimental results for AILearn on various datasets of LivDet database while using LBP, LPQ, BSIF and ResNet-50 features\footnote{Note that Sagem dataset has 1036 spoof images in the test set and Digital-Persona dataset has 1004 live images in the train set. For LBP, LPQ and BSIF features we have considered the original quantity of images, but for ResNet, we have considered 1000 images from each category.}. 
Table \ref{2011} represents the average of AILearn's stability-plasticity values for all combinations of LivDet2011 \cite{DBLP:conf/icb/YambayGDMRS12} datasets as described in Table \ref{partition}. In Table \ref{2011}, we report the average performance of all features for a particular sensor (e.g., Biometrika, DigitalPersona, etc.). The feature-level comparison for individual sensors is given in Figure 3. As a baseline for comparison, we report the performance of the learning model while retraining it using the entire data (e.g., Bio-RT). As we retrain the model in the second phase of learning, the performance values for the first phase are same as without retraining. 

While using LBP features on Biometrika, the average performance degradation for Known Fake (KF) from the first phase to the second phase is $3.34\%$, whereas the performance improvement for New Fake (NF) is $29.75\%$. On LPQ features, the average performance degradation for KF from the first phase to the second phase is $5.57\%$, whereas the performance improvement for NF is $28.03\%$. AILearn performs reasonably well on \textbf{LivDet2011} \cite{DBLP:conf/icb/YambayGDMRS12} Biometrika dataset using ResNet-50 features, with which it yields $57.23\%$ performance improvement from the first phase to the second phase. Also, it is evident that on Biometrika dataset, the best local feature is BSIF, which yields only $1.86\%$ performance degradation on KF whereas $32.69\%$ performance improvement on NF. 

On DigitalPersona dataset while using LBP features, the performance of the model on KF is increased by $8.97\%$ in the second phase, the performance on NF is improved by $15.8\%$ in the second phase. The results of the LPQ feature are significantly well on KF. There is no performance degradation while moving to the second phase; rather, the performance is improved by $2.65\%$. Also, the performance is improved by $39.86\%$ on NF in the second phase. The results on the BSIF feature are adequate as the performance degradation on KF is only $7.05\%$, and the performance improvement is $34.83\%$ with decent overall accuracy. While using ResNet-50 features on DigitalPersona dataset, the performance is improved on KF by $3.19\%$, whereas on NF in the second phase, it is improved by $32.31\%$. 
	
On ItalData dataset, using LBP features, the model yields satisfying stability and reasonable plasticity but the accuracy on NF is not adequate. Using LPQ features the performance degradation on KF is only $2.59\%$, and the performance improvement on NF is $6.55\%$. Using BSIF features, we get outstanding results with only $3.71\%$ performance degradation on KF but $56.37\%$ performance improvement on NF. While using ResNet-50 features on ItalData 2011 dataset, the performance drop on KF from the first phase to the second phase is $4.99\%$, whereas the performance improvement on NF is $2.49\%$. 
	
On Sagem dataset while using LBP features, we get slightly higher performance degradation ($13.71\%$) on KF, which affects the stability of the model, but the plasticity is significantly well with $49.88\%$ performance improvement on NF. While using LPQ features, we get $7.71\%$ performance improvement on KF with $92.67\%$ average accuracy and $47.30\%$ improvement on NF, which results in sound plasticity. The best results we achieve while using BSIF features. We get $3.84\%$ improvement on KF and $50.45\%$ improvement on NF, which results in high plasticity with no compromise on stability. While using ResNet-50 features, the performance of AILearn on KF increases by $5.88\%$ in the second phase, and on NF the performance is increased by $45.99\%$. 
	
We test the performance of AILearn on some samples of \textbf{LivDet2013} \cite{DBLP:conf/icb/GhianiYMTMRS13} and \textbf{LivDet2015} \cite{DBLP:conf/btas/MuraGMRYS15} datasets. On LivDet2013 Biometrika dataset, using BSIF features, we get a performance improvement of $2.87\%$ on NF with adequate overall performance. Using LPQ features, we get an improvement of $11.95\%$ on NF in the second phase. Using ResNet-50 features on LivDet2013 Biometrika dataset, we get an improvement by $187.68\%$ on NF with $99.02\%$ average performance on NF in the second phase. 
    
On LivDet2013 ItalData dataset, while using BSIF feature, the performance improvement is not significant, but the overall performance is reasonably well. While using LPQ features, we get an improvement of $2.45\%$ with an excellent overall performance. While using ResNet-50 features, we get $34.38\%$ increase in the performance on NF in the second phase. 
    
On LivDet2015 Biometrika dataset, using BSIF features, we get an increase of $4.47\%$ on NF in the second phase with an excellent overall performance. Using LPQ features, the performance is increased by $34.11\%$ in the second phase. While using ResNet features, the performance is increased by $35.56\%$ on NF in the second phase.  On LivDet2015 DigitalPersona dataset, using BSIf features, there is no significant improvement, but the performance is $90.4\%$ in the second phase. Using LPQ features, there is an increase of $13.07\%$ on NF in the second phase. Using ResNet-50 features, we get an increase of $4.25\%$ on NF in the second phase. 

	\renewcommand{\arraystretch}{1.3}
	\begin{table*}[h]\caption{Stability-Plasticity calculation on LivDet 2011 \cite{DBLP:conf/icb/YambayGDMRS12} dataset}
		\begin{center}
			\resizebox{\textwidth}{!}{\begin{tabular}{|l|c|c|c|c|c|c|c|c|c|c|c|c|}
				\hline
				\multirow{3}{*}{DataSet} & \multicolumn{12}{c|}{AILearn}                         \\ \cline{2-13} 
				& \multicolumn{3}{c|}{I. KF}                                                                                                                                      & \multicolumn{3}{c|}{I. NF}                                                                                                                                        & \multicolumn{3}{c|}{II. KF}                                                                                                                                     & \multicolumn{3}{c|}{II. NF}                                                                                                                                       \\ \cline{2-13} 
				& \begin{tabular}[c]{@{}c@{}}Acc\\ (\%)\end{tabular} & \begin{tabular}[c]{@{}c@{}}BPCER\\ (0-1)\end{tabular} & \begin{tabular}[c]{@{}c@{}}APCER\\ (0-1)\end{tabular} & \begin{tabular}[c]{@{}c@{}}Acc\\ (\%)\end{tabular} & \begin{tabular}[c]{@{}c@{}}BPCER\\ (0-1)\end{tabular} & \begin{tabular}[c]{@{}c@{}}APCER\\ (0-1)\end{tabular} & \begin{tabular}[c]{@{}c@{}}Acc\\ (\%)\end{tabular} & \begin{tabular}[c]{@{}c@{}}BPCER\\ (0-1)\end{tabular} & \begin{tabular}[c]{@{}c@{}}APCER\\ (0-1)\end{tabular} & \begin{tabular}[c]{@{}c@{}}Acc\\ (\%)\end{tabular} & \begin{tabular}[c]{@{}c@{}}BPCER\\ (0-1)\end{tabular} & \begin{tabular}[c]{@{}c@{}}APCER\\ (0-1)\end{tabular} \\ \hline
Bio                           & 80.23 & 0.15  & 0.31  & 55.39 & N/A   & 0.45  & 78.87  & 0.4   & 0.13  & 76.18  & N/A   & 0.23  \\ \hline
Bio-RT                        & 80.23 & 0.15  & 0.31  & 55.39 & N/A   & 0.45  & 84.58  & 0.12  & 0.22  & 67.56  & N/A   & 0.32  \\ \hline
Dig                           & 89.24 & 0.02  & 0.31  & 26.46 & N/A   & 0.73  & 90.88  & 0.07  & 0.11  & 35.02  & N/A   & 0.64  \\ \hline
Dig-RT                        & 89.24 & 0.02  & 0.31  & 26.46 & N/A   & 0.73  & 90.72  & 0.02  & 0.26  & 33.51  & N/A   & 0.66  \\ \hline
Ital                          & 80.45 & 0.11  & 0.37  & 49.35 & N/A   & 0.51  & 78.16  & 0.18  & 0.31  & 56.7   & N/A   & 0.43  \\ \hline
Ital-RT                       & 80.45 & 0.11  & 0.37  & 49.35 & N/A   & 0.51  & 81.38  & 0.11  & 0.36  & 52.22  & N/A   & 0.46  \\ \hline
Sag                           & 81.3  & 0.14  & 0.29  & 35.62 & N/A   & 0.64  & 83.08  & 0.22  & 0.04  & 52.96  & N/A   & 0.47  \\ \hline
Sag-RT                        & 81.3  & 0.14  & 0.29  & 35.62 & N/A   & 0.64  & 84.31  & 0.14  & 0.19  & 42.23  & N/A   & 0.56  \\ \hline
\end{tabular}}
		\end{center}\label{2011}
	\end{table*}	
	
\renewcommand{\arraystretch}{1.3}
\begin{table*}[h]\caption{Stability-Plasticity calculation on LivDet 2013 \cite{DBLP:conf/icb/GhianiYMTMRS13}-LivDet 2015 \cite{DBLP:conf/btas/MuraGMRYS15} dataset}
		\begin{center}
			\resizebox{\textwidth}{!}{\begin{tabular}{|l|c|c|c|c|c|c|c|c|c|c|c|c|}
				\hline
				\multirow{3}{*}{DataSet} & \multicolumn{12}{c|}{AILearn}                         \\ \cline{2-13} 
				& \multicolumn{3}{c|}{I. KF}                                                                                                                                      & \multicolumn{3}{c|}{I. NF}                                                                                                                                        & \multicolumn{3}{c|}{II. KF}                                                                                                                                     & \multicolumn{3}{c|}{II. NF}                                                                                                                                       \\ \cline{2-13} 
				& \begin{tabular}[c]{@{}c@{}}Acc\\ (\%)\end{tabular} & \begin{tabular}[c]{@{}c@{}}BPCER\\ (0-1)\end{tabular} & \begin{tabular}[c]{@{}c@{}}APCER\\ (0-1)\end{tabular} & \begin{tabular}[c]{@{}c@{}}Acc\\ (\%)\end{tabular} & \begin{tabular}[c]{@{}c@{}}BPCER\\ (0-1)\end{tabular} & \begin{tabular}[c]{@{}c@{}}APCER\\ (0-1)\end{tabular} & \begin{tabular}[c]{@{}c@{}}Acc\\ (\%)\end{tabular} & \begin{tabular}[c]{@{}c@{}}BPCER\\ (0-1)\end{tabular} & \begin{tabular}[c]{@{}c@{}}APCER\\ (0-1)\end{tabular} & \begin{tabular}[c]{@{}c@{}}Acc\\ (\%)\end{tabular} & \begin{tabular}[c]{@{}c@{}}BPCER\\ (0-1)\end{tabular} & \begin{tabular}[c]{@{}c@{}}APCER\\ (0-1)\end{tabular} \\ \hline

2013-Bio                      & 98.03 & 0.02  & 0.02  & 93.82 & N/A   & 0.06  & 97.6   & 0.03  & 0     & 96.52  & N/A   & 0.03  \\ \hline
2013-Ital                     & 97.8  & 0     & 0.07  & 84.23 & N/A   & 0.16  & 95.10  & 0     & 0.07  & 84.25  & N/A   & 0.16  \\ \hline
2015-Bio                      & 83.26 & 0.15  & 0.03  & 89.4  & N/A   & 0.11  & 81.34  & 0.27  & 0     & 93.4   & N/A   & 0.07  \\ \hline
2015-Dig                      & 83.64 & 0.22  & 0.06  & 89.37 & N/A   & 0.11  & 81.33  & 0.25  & 0.05  & 90.4   & N/A   & 0.1   \\ \hline
\end{tabular}} 
	
		\end{center}\label{2013-15}
	\end{table*}	
\subsection{Feature-level comparison}\label{feature}
\begin{center}
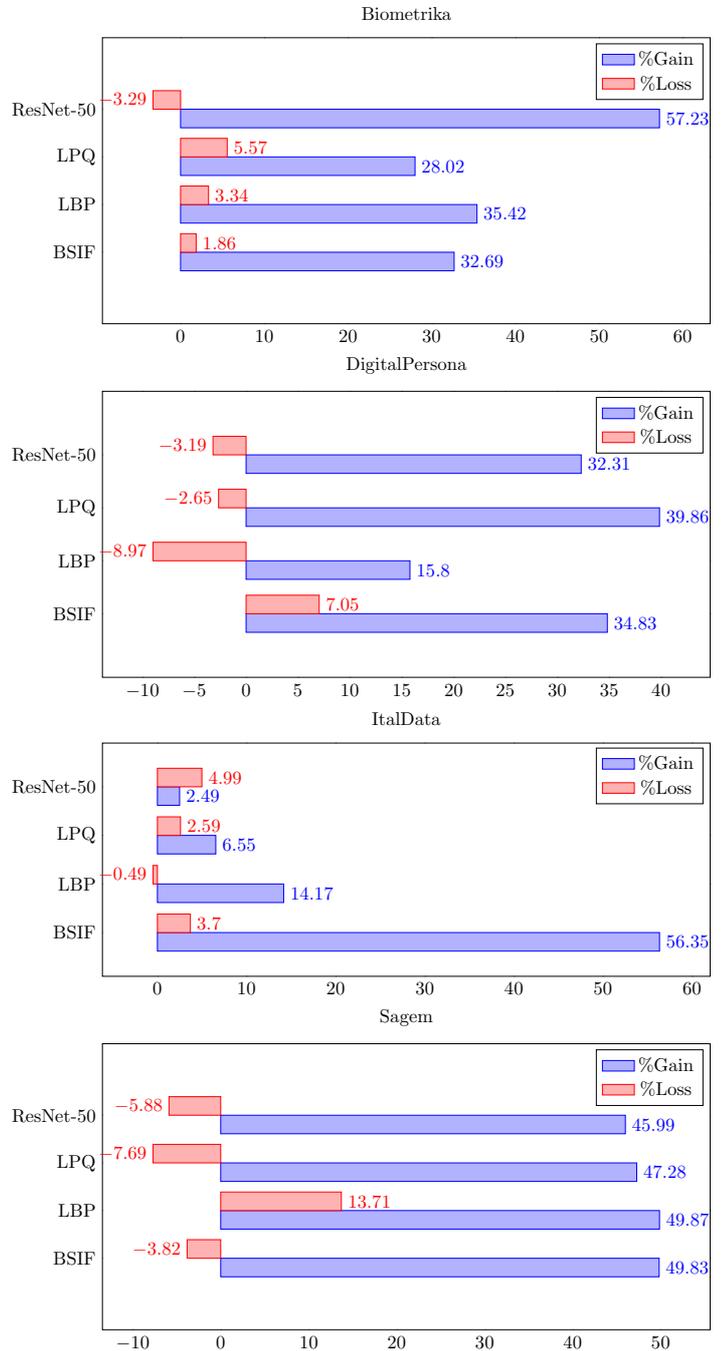
\begin{figure}[]
\begin{subfigure}{.4\textwidth}
\begin{tikzpicture}[scale=0.7]
\centering
  \begin{axis}[
    title    = Biometrika,
    height=7cm,
      width=13cm,
      xbar=0pt,
      bar width=1em,
      legend style={xshift=0.1cm},
      ytick=data,
      enlarge x limits  = 0.1,
      enlarge y limits  = 0.5,
      tick align=inside,
      tickwidth = 0pt,
      symbolic y coords = {%
        {BSIF}, {LBP}, {LPQ}, {ResNet-50}},
        nodes near coords,
      nodes near coords align={horizontal},
    ]
    \addplot +[area legend] coordinates { (32.69,{BSIF}) (35.42,{LBP})
           (28.02,{LPQ}) (57.23,{ResNet-50}) };
    \addplot +[area legend] coordinates { (1.86,{BSIF}) (3.34,{LBP})
           (5.57,{LPQ})  (-3.29,{ResNet-50})  };
    
    \legend{\%Gain, \%Loss}
\end{axis}

\end{tikzpicture}
\end{subfigure}

    \begin{subfigure}{.3\textwidth}
\begin{tikzpicture}[scale=0.7]
 \centering 
  \begin{axis}[
    title    = DigitalPersona,
    height=7cm,
      width=13cm,
      xbar=0pt,
      bar width=1em,
      legend style={xshift=0.1cm},
      ytick=data,
      enlarge x limits  = 0.1,
      enlarge y limits  = 0.4,
      tick align=inside,
      tickwidth = 0pt,
      symbolic y coords = {%
        {BSIF}, {LBP}, {LPQ}, {ResNet-50}},
        nodes near coords,
      nodes near coords align={horizontal},
    ]
    \addplot +[area legend] coordinates { (34.83,{BSIF}) (15.8,{LBP})
           (39.86,{LPQ}) (32.31,{ResNet-50}) };
    \addplot +[area legend] coordinates {(7.05,{BSIF}) (-8.97,{LBP})
           (-2.65,{LPQ})  (-3.19,{ResNet-50})  };
    \legend{\%Gain, \%Loss}
\end{axis}
\end{tikzpicture}
\end{subfigure} 

\begin{subfigure}{.3\textwidth}
\begin{tikzpicture}[scale=0.7]
  \centering
  \begin{axis}[
    title    = ItalData,
    height=6cm,
      width=13cm,
      xbar=0pt,
      bar width=1em,
      legend style={xshift=0.1cm},
      ytick=data,
      enlarge x limits  = 0.1,
      enlarge y limits  = 0.3,
      tick align=inside,
      tickwidth = 0pt,
      symbolic y coords = {%
        {BSIF}, {LBP}, {LPQ}, {ResNet-50}},
        nodes near coords,
      nodes near coords align={horizontal},
    ]
    \addplot +[area legend] coordinates { (56.35,{BSIF}) (14.17,{LBP})
           (6.55,{LPQ}) (2.49,{ResNet-50}) };
    \addplot +[area legend] coordinates {(3.7,{BSIF}) (-0.49,{LBP})
           (2.59,{LPQ})  (4.99,{ResNet-50})  };
    \legend{\%Gain, \%Loss}
\end{axis}
\end{tikzpicture}
\end{subfigure}

\begin{subfigure}{.3\textwidth}
\begin{tikzpicture}[scale=0.7]
  \centering
  \begin{axis}[
    title    = Sagem,
    height=7cm,
      width=13cm,
      xbar=0pt,
      bar width=1em,
      legend style={xshift=0.1cm},
      ytick=data,
      enlarge x limits  = 0.1,
      enlarge y limits  = 0.5,
      tick align=inside,
      tickwidth = 0pt,
      symbolic y coords = {%
        {BSIF}, {LBP}, {LPQ}, {ResNet-50}},
        nodes near coords,
      nodes near coords align={horizontal},
    ]
    \addplot +[area legend] coordinates { (49.83,{BSIF}) (49.87,{LBP})
           (47.28,{LPQ}) (45.99,{ResNet-50})  };
    \addplot +[area legend] coordinates {(-3.82,{BSIF}) (13.71,{LBP})
           (-7.69,{LPQ})  (-5.88,{ResNet-50})  };
    \legend{\%Gain, \%Loss}
\end{axis}

\end{tikzpicture}
\end{subfigure}\caption{Comparison of the performance of AILearn when used with different features shown on Y axis. Percentage Gain on NF and percentage Loss on KF while learning in second phase are shown on X axis.}\label{fig3}
\end{figure}
\end{center}
Figure \ref{fig3} represents the comparison among various features used with AILearn on LivDet 2011 \cite{DBLP:conf/icb/YambayGDMRS12} datasets. We emphasise on the percentage gain on NF and percentage loss on KF in subsequent phases. Ideally, the features on which the percentage gain is high and percentage loss is low are the most suited for application. From the Figure \ref{fig3}, we can see on Biometrika, using ResNet features we get a performance gain of $57.23\%$, whereas the loss in performance on KF is $-3.29\%$. The $'-'$ symbol represents that instead of performance loss on KF, we observe a gain of $-3.29\%$. Among the handcrafted features, LBP yields the highest percentage gain and BSIF yields the lowest percentage loss.

On Digital Persona dataset, the highest performance gain is observed using LPQ features and the lowest performance loss is observed using LBP features. On an average, performance of LPQ features is the most adequate on this dataset. 

On Ital Data dataset, we observe the highest performance gain using BSIF features and the lowest performance loss by using LBP features. Overall, the performance of BSIF features is the most adequate.

On Sagem dataset, we observe the highest gain while using LBP features and the lowest loss while using LPQ features. The performance of ResNet is close to LPQ, but on an average LPQ is the most suited for this dataset. \textbf{Similar observations has been obtained for the same features on LivDet 2013 \cite{DBLP:conf/icb/GhianiYMTMRS13} and LivDet 2015 \cite{DBLP:conf/btas/MuraGMRYS15}, not included for the sake of space}.   
\subsection{Comparison with State-of-the-art}
In this section, we compare the performance of AILearn with the current state-of-the-art. To evaluate the performance of AILearn along with the existing work in incremental setting, we compare BPCER of AILearn on New Fake (NF) and Known Fake (KF) mentioned in the tables given in Section \ref{results} with \textit{ferrfake when ferrlive=10\%} from \cite{KHO201952} and the results of \cite{7180344} mentioned in the paper. The comparison results on LivDet2011 datasets are given in Table \ref{sota}. As can be seen from the Table \ref{sota}, performance is evaluated based on two metrics: percentage loss in FPR (or \textit{ferrfake}) on NF and percentage change in FPR on KF. We emphasize that for an incremental learning algorithm, it is essential to have a decent percentage loss in FPR on NF, and the percentage change in FPR on KF must always be negative. As AILearn makes use of three hand-crafted features and one type of deep features, we report the best performance yielded by any type of feature (indicated in parentheses). It is evident from the Table \ref{sota}, that none of the state-of-the-art models produce all negative values in percentage change in FPR on KF, but AILearn does that while maintaining a good percentage loss in FPR on NF. \textbf{Similar observations have been noted for LivDet 2013 \cite{DBLP:conf/icb/GhianiYMTMRS13} and LivDet 2015 \cite{DBLP:conf/btas/MuraGMRYS15}, not included for the sake of space}.
\renewcommand{\arraystretch}{1.3}
\begin{table*}[h]\caption{Performance evaluation of AILearn in comparison to the state-of-the-art \cite{KHO201952, 7180344} on LivDet2011 \cite{DBLP:conf/icb/YambayGDMRS12} datasets. In this table, FPR, NF and KF denotes false positive rate, new fake and known false, respectively.}
	\centering
	\resizebox{\textwidth}{!}{\begin{tabular}{|l|c|c|c|c|c|c|c|c|}
		\hline
		\multirow{2}{*}{Dataset} & \multicolumn{2}{c|}{Kho. et. al. \cite{KHO201952}}                                                                                                  & \multicolumn{2}{c|}{Rattani et. al. Feature level \cite{7180344} }                                                                                              & \multicolumn{2}{c|}{Rattani et. al. Score level \cite{7180344}}                                                                                                & \multicolumn{2}{c|}{AILearn}                                                                                                            \\ \cline{2-9} 
		& \begin{tabular}[c]{@{}c@{}}\%loss in \\ \\ FPR on NF\end{tabular} & \begin{tabular}[c]{@{}c@{}}\%change in \\ \\ FPR on KF\end{tabular} & \begin{tabular}[c]{@{}c@{}}\%loss in \\ \\ FPR on NF\end{tabular} & \begin{tabular}[c]{@{}c@{}}\%change in \\ \\ FPR on KF\end{tabular} & \begin{tabular}[c]{@{}c@{}}\%loss in \\ \\ FPR on NF\end{tabular} & \begin{tabular}[c]{@{}c@{}}\%change in \\ \\ FPR on KF\end{tabular} & \begin{tabular}[c]{@{}c@{}}\%loss in \\ \\ FPR on NF\end{tabular} & \begin{tabular}[c]{@{}c@{}}\%change in \\ \\ FPR on KF\end{tabular} \\ \hline
		Biometrika               & 62.72                                                             & -10.50                                                              & 64.20                                                             & 4.44                                                                & 70.27                                                             & -25.48                                                              & 53.85(ResNet)                                                     & -74.36                                                              \\ \hline
		DigitalPersona           & 93.95                                                             & 57.94                                                               & 89.15                                                             & 37.78                                                               & 72.20                                                             & 91.29                                                               & 25.86(BSIF)                                                       & -57.14                                                              \\ \hline
		ItalData                 & 67.90                                                             & -7.54                                                               & 55.55                                                             & 3.18                                                                & 39.32                                                             & 19.88                                                               & 60.00(ResNet)                                                     & -100                                                                \\ \hline
		Sagem                    & 89.49                                                             & 44.61                                                               & 89.49                                                             & 39.66                                                               & 89.09                                                             & 13.87                                                               & 60.00(LBP)                                                        & -86.67                                                              \\ \hline
	\end{tabular}}	
\label{sota}
\end{table*}
\subsection{Discussion on results}
	We describe the results for AILearn by the Tables \ref{2011}-\ref{2013-15}. We conduct the experiments with the motivation of exploiting the incremental ability of AILearn. For that, our emphasis is on reporting the stability and plasticity of the proposed model. We highlight the performance degradation on the known spoof fingerprints and the performance improvement on the new spoof fingerprints before and after learning the new fake data. Our experimental results justify our motivation as we are able to achieve high plasticity with no or negligible loss in stability. As a baseline, we consider a model which requires to be retrained using the entire data in the second learning phase. Ideally, such a model must not encounter performance drop on KF in the second phase and then there must be a significant performance improvement on NF. As we compare the results of AILearn with and without retraining the model with entire data, our results without retraining seem to be satisfying the motivation of the paper. 
    
    In some cases, we get better stability but lesser plasticity while using a particular feature and vice-versa. In some cases, we get reasonable stability and plasticity, but the overall accuracy of the model is not adequate with that feature. This kind of results supports the famous no-free-lunch theorem \cite{585893}, which states that no one model works best for every problem. Therefore, it is advised in machine learning to try multiple models with different settings and find one that works best for a particular problem. On an average, AILearn performs reasonably well on all the datasets with high performance gain on NF and low/negligible performance loss on KF. In addition, the overall accuracy of the proposed model is also high, with highest accuracy approaching $96.52\%$ on NF. 
   \subsubsection{Performance of AILearn in comparison with baseline and state-of-the-art}
   AILearn performs well both in respect of stability and plasticity. Table \ref{2011} shows that AILearn yield better plasticity than the baseline where we retrain the model from the scratch. AILearn without retraining the model gives better stability in two of the four cases of LivDet2011 \cite{DBLP:conf/icb/YambayGDMRS12} datasets, which encourages the motivation for incremental learning. In addition, the performance on LivDet2013 \cite{DBLP:conf/icb/GhianiYMTMRS13} and LivDet2015 \cite{DBLP:conf/btas/MuraGMRYS15} is reasonable well with the highest accuracy reaching $96.52\%$ on NF and $97.6\%$ on KF. 
   \subsubsection{Performance of AILearn using various features}
   As described in Section \ref{feature}, AILearn performs well on every type of features. This feature level comparison provides useful insights on the performance of handcrafted features and deep features. Most often, it is argued that the deep features outperform handcrafted features in almost every case, therefore handcrafted features must be discarded. On contrary, this study proves that handcrafted features give neck to neck competition and in some cases outperform the deep features. Therefore, a single type of features can not be trusted to perform well in every case.  
    
    \section{Conclusions}\label{conclusions}
	An incremental learning algorithm must be able to learn from the newly added data while retaining the already acquired knowledge from the past data. We propose a novel incremental learning model AILearn and show its working mechanism on spoof fingerprint detection. The proposed algorithm is able to learn the new spoof fingerprints from the current learning phase while maintaining its performance on the ``live", and ``spoof" fingerprints learned in the previous learning phase. AILearn is an adaptive way of learning as it adapts to the similarity inherently present in the data. Also, AILearn is an efficient algorithm as it discards the already seen data and keeps only knowledge extracted from it. By doing so, space can be reused for storing the upcoming data. With this motivation, we conducted our experiments and proved the efficacy of AILearn. We highlight the stability and plasticity features of AILearn on LivDet2011, LivDet2013 and LivDet2015 dataset and conclude that the proposed framework improves its performance in the new learning phase by $49.57\%$ on an average, without considerable degradation in the existing knowledge. This study provides critical insights into feature level comparison. We provide a detailed comparison of handcrafted features v/s deep features. As a part of future work, integration of the hand-crafted and deep features in AILearn will be investigated at feature and score level \cite{7180344} for further performance enhancement. 
	\bibliographystyle{elsarticle-num}
	\bibliography{AILearn}
	
	%
	
	
	
	
	
	

\end{document}